\documentclass[sigconf]{acmart}

\AtBeginDocument{%
  \providecommand\BibTeX{{%
    \normalfont B\kern-0.5em{\scshape i\kern-0.25em b}\kern-0.8em\TeX}}}

\setcopyright{acmcopyright}
\copyrightyear{2024}
\acmYear{2024}
\acmDOI{XXXXXXX.XXXXXXX}

\acmConference[XX]{Anonymous Conference 2024}{XX XX--XX,
  2024}{XX}
\acmPrice{15.00}
\acmISBN{978-1-4503-XXXX-X/18/06}

\usepackage{microtype}
\usepackage{mathtools}

\usepackage{amsthm}

\usepackage{subfigure}  
\usepackage{algorithmicx}  
\usepackage{algpseudocode}  
\usepackage{amsmath}
\usepackage{multicol}
\usepackage{multirow}
\usepackage{xspace}
\usepackage{svg}
\usepackage{xcolor}

\begin{document}
\newcommand{\name}[0]{GTPool\xspace}
\newcommand{\vname}[0]{GTPool-V\xspace}

\makeatletter
\DeclareRobustCommand\onedot{\futurelet\@let@token\@onedot}
\def\@onedot{\ifx\@let@token.\else.\null\fi\xspace}
\def\eg{\emph{e.g}\onedot} \def\Eg{\emph{E.g}\onedot}
\def\ie{\emph{i.e}\onedot} \def\Ie{\emph{I.e}\onedot}
\def\cf{\emph{c.f}\onedot} \def\Cf{\emph{C.f}\onedot}
\def\etc{\emph{etc}\onedot} \def\vs{\emph{vs}\onedot}
\def\wrt{w.r.t\onedot} \def\dof{d.o.f\onedot}
\def\etal{\emph{et al}\onedot}



\title{Diversified Node Sampling based Hierarchical Transformer Pooling for Graph Representation Learning}



\author{Gaichao Li\textsuperscript{1,2,3}, Jinsong Chen$^{1,2,3}$, John E. Hopcroft$^{4}$, Kun He$^{2,3,\dag}$}

\thanks{$\dag$ Corresponding author.}
\affiliation{%
  \institution{$^{1}$ Institute of Artificial Intelligence, Huazhong University of Science and Technology
  \city{Wuhan}
  \country{China}}
  \institution{$^{2}$ School of Computer Science and Technology, Huazhong University of Science and Technology   
  \city{Wuhan}
  \country{China}}
  \institution{$^{3}$ Hopcroft Center on Computing Science, Huazhong University of Science and Technology   
  \city{Wuhan}
  \country{China}}
  \institution{$^{4}$ Department of Computer Science, Cornell University   
  \city{Ithaca}
  \country{USA}}
}
\email{{gaichaolee,chenjinsong}@hust.edu.cn, jeh@cs.cornell.edu,brooklet60@hust.edu.cn}

\def\authors{Gaichao Li, Jinsong Chen, John E. Hopcroft, Kun He}

\renewcommand{\shortauthors}{Gaichao, et al.}

\begin{abstract}
Graph pooling methods have been widely used on downsampling graphs, achieving impressive results on multiple graph-level tasks like graph classification and graph generation.
An important line called node dropping pooling aims at exploiting learnable scoring functions to drop nodes with comparatively lower significance scores. 
However, existing node dropping methods suffer from two limitations: (1) for each pooled node, these models struggle to capture long-range dependencies since they mainly take GNNs as the backbones; (2) pooling only the highest-scoring nodes tends to preserve similar nodes, thus discarding the affluent information of low-scoring nodes. 
To address these issues, we propose a Graph Transformer Pooling method termed \name, which introduces Transformer to node dropping pooling to efficiently capture long-range pairwise interactions and meanwhile sample nodes diversely.
Specifically, we design a scoring module based on the self-attention mechanism that takes both global context and local context into consideration, measuring the importance of nodes more comprehensively. 
\name further utilizes a diversified sampling method named Roulette Wheel Sampling (RWS) that is able to flexibly preserve nodes across different scoring intervals instead of only higher scoring nodes.
In this way, \name could effectively obtain long-range information and select more representative nodes.  
Extensive experiments on 11 benchmark datasets demonstrate the superiority of \name over existing popular graph pooling methods. 
\end{abstract}

\ccsdesc[500]{Computing methodologies~Supervised learning by classification}
\ccsdesc[300]{Mathematics of computing~Graph algorithms}

\keywords{Graph Pooling, Graph Transformer, Node Dropping Pooling, Diversified Node Sampling, Hierarchical Pooling}



\maketitle

\section{Introduction}

Recent years have witnessed an impressive growth in graph representation learning due to its wide application across various domains such as social network~\cite{social}, chemistry~\cite{gmt}, biology~\cite{biology} and recommender systems~\cite{recommend}.
Among numerous techniques for graph representation learning, graph pooling~\cite{diffpool}, a prominent branch aiming to map a set of nodes into a compact form to represent the entire graph, has attracted great attention in the literature. 

Existing graph pooling approaches can be roughly divided into two categories~\cite{pooling_survey}: flat pooling and hierarchical pooling.   
The former attempts to design pooling operations (\eg sum-pool or mean-pool) to generate a graph representation in one step.
The latter conducts the pooling operation on graphs in a hierarchical manner, which is beneficial to preserving more structural information than flat pooling methods.
Due to the merits of hierarchical pooling, it has 
become the leading architecture for graph pooling.

There are mainly two types of hierarchical pooling methods~\cite{pooling_survey}, namely node clustering and node dropping.
Node clustering pooling considers graph pooling as a node clustering problem, which maps the nodes into a set of clusters by learning a cluster assignment matrix.
While node dropping pooling methods mainly focus on learning a scoring function, then utilize the scores to select top-$K$ nodes and drop all other nodes to downsample the graph. 
Compared with node clustering pooling, node dropping pooling is usually more efficient and scalable, especially on large graphs.
Recently, powerful node dropping pooling methods, including SAGPool~\cite{sagpool}, TopKPool~\cite{topkpool} and ASAP~\cite{asap}, have been developed and achieved good performance in learning the graph representations.
These methods share a similar scheme: identifying and sampling important nodes to build the coarsened graphs hierarchically. 

Despite effectiveness, we observe that existing node dropping pooling methods share two common weaknesses: overlooking long-range dependencies and inflexible sampling strategy.

\textbf{\textit{Overlooking long-range dependencies.}}
Existing node dropping methods leverage GNN layers as the backbone, which is naturally inefficient in capturing long-range dependencies of the input graph.
Though there are several graph pooling methods~\cite{graphtrans,gmt} that introduce the Transformer architecture to obtain long-range information, they obey the flat pooling design, restricting their capacity to learn the hierarchical graph structure. 
Hence, there still lack methods that could effectively model long-range relationships meanwhile encoding hierarchical graph structures.

\textbf{\textit{Inflexible sampling strategy.}}
Most existing node dropping methods additionally suffer from only pooling the top-$K$ nodes with higher scores and discarding all other lower scoring nodes.  As described in \cite{reppool}, connected nodes may have similar scores, as their node features tend to be similar. 
Therefore, preserving only $K$ higher scoring nodes may results in throwing away the whole neighborhood of nodes and inclines to highlight similar nodes.
To mitigate this issue, except for the importance score, RepPool~\cite{reppool} extra adopts a representativeness score to help select nodes from different substructures. Nevertheless, 
as a top-$K$ sampling method, it is still hard 
to ensure the diversity of the pooled nodes.       
As a result, existing node dropping pooling methods can only generate suboptimal graph representations as they ignore the diversities of the sampled nodes.

To tackle the above 
issues, we propose a novel graph pooling 
method, termed Graph Transformer Pooling (\name), which incorporates the Transformer architecture into node dropping pooling to help the pooled nodes capture long-range information, meanwhile we develop a new node sampling method that can diversely select nodes across different scoring segments.
Specifically, we first construct a node scoring module based on the attention matrix to evaluate the significance of nodes based on both global and local context. 
Then, we utilize a plug-and-play and parameter-free sampling module called Roulette
Wheel Sampling (RWS) 
that is capable of diversely sampling nodes from different scoring intervals.
With the indices of sampled nodes, the attention matrix can be refined from $n \times n$ to $\lceil \mu n \rceil \times n$, where $n$ is the number of nodes and  $\mu$ is the pooling ratio.
In this way, each pooled node can efficiently obtain information from all nodes on the original graph.
Besides, our \name layer can be integrated with GNN layer to learn graph structures in a hierarchical manner.

To investigate the effectiveness of \name, we further conduct experiments on 11 popular benchmarks for the graph classification task.
Empirical results show that our \name consistently outperforms or achieves competitive performance compared with many state-of-the-art pooling methods. 

Our main contributions are summarized as follows:

\begin{itemize}
    \item 
    To our knowledge, this is the first work that introduces Transformer to node dropping pooling. And we further propose a novel graph pooling method called \name, which can be combined with GNN layers to perform hierarchical pooling and makes the pooled nodes capable of capturing long-range relationships. 
    \item We design a plug-and-play and parameter-free sampling strategy called Roulette Wheel Sampling (RWS) that can flexibly select representative nodes across all scoring intervals. 
    Moreover, RWS can boost the performance of other top-$K$ selection-based graph pooling methods by directly replacing the sampling strategy.

    \item 
    Extensive comparisons on 11 datasets with SOTA baseline methods validate the superior performance of our \name on the graph classification task. 
\end{itemize}





\section{Related Work} \label{RW}
\begin{figure*}
  \includegraphics[width=12.5cm]{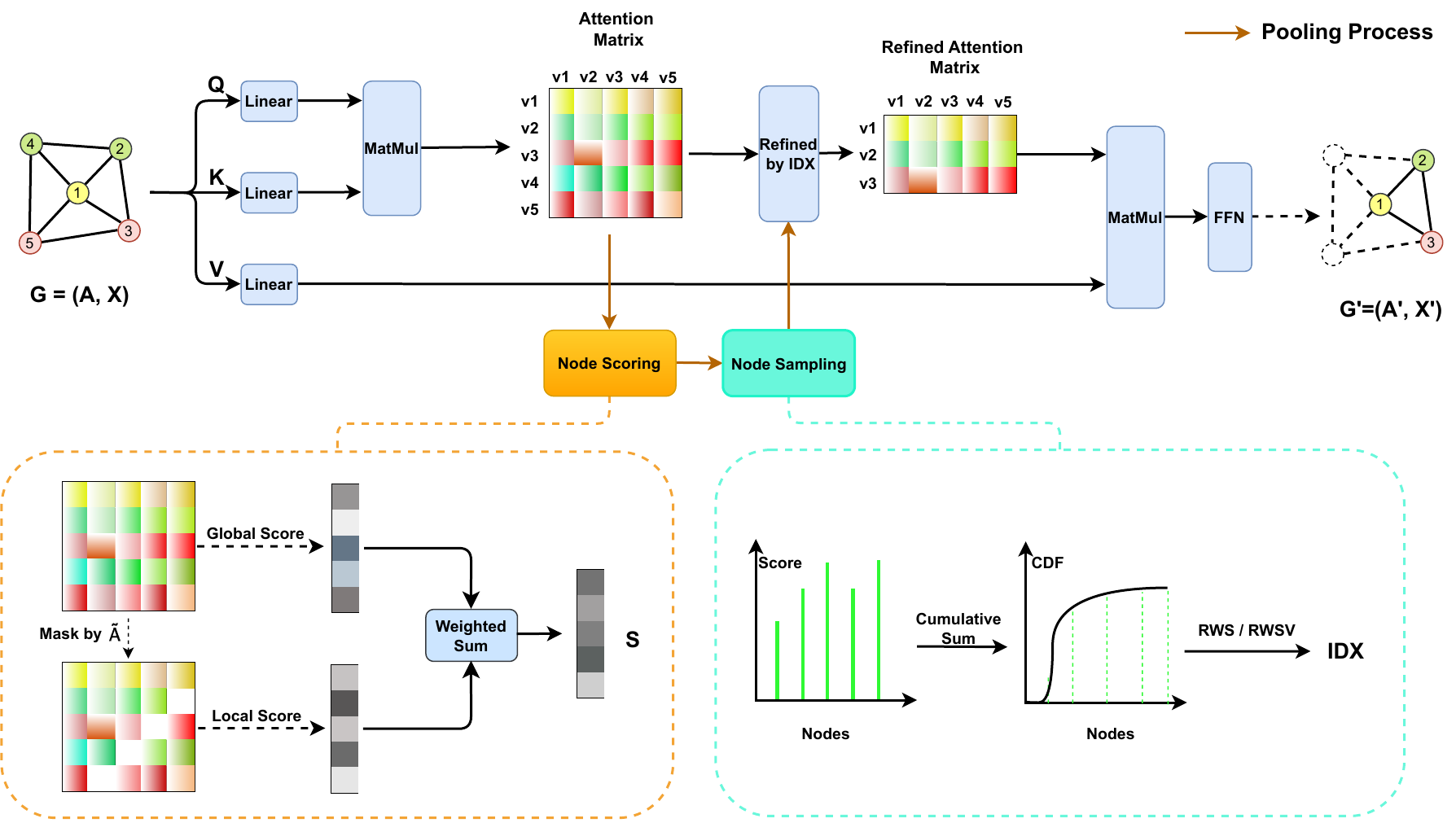}
  \caption{An illustration of the proposed \name layer. The scoring module first utilizes the attention matrix to evaluate the significance of nodes from both the local and global views. 
  Then, the sampling module leverages the significance score to diversely select nodes from different sampling intervals. And the attention matrix can be refined with the indices of the sampled nodes. Hence, the pooled nodes can efficiently capture long-range interactions from all other nodes on the original graph. Here we show single head attention and omit the residual connection for better illustration.}
  \label{fig:frame}
\end{figure*}

In this section, we first review the classical graph neural networks and node dropping pooling methods, then briefly present the most recent 
emerging works that introduce Transformer to graph pooling.

\textbf{Graph Neural Networks. }
Existing graph neural networks (GNNs) can be generally classified into spectral and spatial models. 
Based on the graph spectral theory, spectral models usually perform convolutional operations in the Fourier domain. 
Bruna~\etal~\cite{spectral} first propose the graph convolutional operation by adopting spectral filters, but it cannot scale to large graphs due to the high-computational cost and non-locality property.
To improve the efficiency, Defferrard~\etal~\cite{defferrard} utilize an approximation of $K$-polynomial filters that are strictly localized within the radius $K$. Kipf and Welling~\cite{gcn} further design a simplified model by leveraging the first-order approximation of the Chebyshev expansion. 
Different from spectral models, the spatial approaches can directly work on graphs by aggregating information from the local neighbors to update the central node.
Among them, GAT~\cite{gat} employs the attention mechanism in Transformer to distinguish the importance of information from different nodes. 
As another typical work, GraphSage~\cite{graphsage} provides a general inductive framework that can generate embeddings for unseen nodes by sampling and aggregating features from the local neighborhood of each node.
Till now, a variety of GNNs~\cite{appnp,rlp,gcnii,glognn} have been proposed to handle different kinds of graphs.
Nevertheless, these GNNs belong to flat methods that focus on learning node representations for all nodes, but hard to work on the graph classification task directly.
Besides, GNNs also suffer from over-smoothing~\cite{oversmoothing} and over-squashing~\cite{oversq} problems, causing them unable to efficiently capture long-range pairwise dependencies. 

\textbf{Node Dropping Pooling. }
Compared to flat pooling, hierarchical pooling takes the hierarchical structures of graphs into consideration, therefore generating more expressive graph representations.
With different coarsening manners, hierarchical pooling can be roughly grouped into two categories: node clustering pooling and node dropping pooling. 
Node clustering pooling~\cite{diffpool, mincutpool} casts the graph pooling problem into the node clustering problem that map the nodes into a bunch of clusters. 
As another important line, the key procedure of node dropping pooling~\cite{sagpool, topkpool, asap} is to adopt learnable scoring functions to drop nodes with comparatively lower scores. 
For instance, Gao~\etal~\cite{topkpool} propose a simple method named TopKPool that utilizes a learnable vector to calculate the projection scores and uses the scores to select the top ranked nodes.
Similarly, as an important variant of TopKPool, SAGPool~\cite{sagpool} leverages a GNN layer to provide self-attention scores that both consider attributed and topological information and also sample the higher scoring nodes.
Different from the above two methods, ASAP~\cite{asap} adopts a novel self-attention network along with a modified GNN formulation to capture the importance of each node.
Subsequently, multiple types of node dropping pooling methods have emerged by designing more sophisticated scoring functions and more reasonable graph coarsening to select more representative nodes and to retain more important structural information, respectively. 

However, based on existing node dropping methods, we find the obtained pooled nodes struggle to capture long-range information, restricting the expressiveness of graph representations.
Besides, these methods also suffer from solely sampling higher scoring nodes with similar features, which may results in performing less satisfactory when multiple dissimilar nodes contribute to the graph representations. 





\textbf{Graph Pooling with Transformer. }
Transformer-based models~\cite{graphormer,sat,nagphormer} have shown impressive performance on various graph mining tasks.
Hence, several works have attempted to introduce the Transformer architecture to the graph pooling domain.
GraphTrans~\cite{graphtrans} employs a special learnable token to aggregate all pairwise interactions into a single classification vector, which can be actually regarded as a kind of flat pooling.
Jinheon~\etal~\cite{gmt} formulate the graph pooling problem as a multiset encoding problem and propose GMT which is a multi-head attention based global pooling layer.
To sum up, there still lacks an effective method that can combine Transformer with graph pooling problem and meanwhile perform hierarchical pooling to obtain more accurate graph representations. 

In this paper, we propose \name that attempts to introduce Transformer to node dropping pooling methods. 
\name not only can pool graphs in a hierarchical way, but it also samples nodes in a diverse way, enabling it to generate better graph representations. 

\section{Preliminaries}
For preliminaries, we first present the notations and terminologies of attributed graphs for problem formulation, then shortly recap the general descriptions of graph neural networks and the classic Transformer architecture.

\textbf{Notations. }
In this paper, we concentrate on the graph classification task, which aims at mapping each of the graphs to a set of labels. 
We define an arbitrary attributed graph as $\mathcal{G}=(\mathcal{V}, \mathcal{E}, \mathbf{X})$, 
where $\mathcal{V}$ and $\mathcal{E}$ are the node set and edge set, respectively.  
Besides, we have $\mathbf{A} \in \mathbb{R}^{n \times n}$ represents the adjacency matrix, and $\mathbf{X} \in \mathbb{R}^{n \times d}$ represents the feature matrix of $\mathcal{G}$ in which $n$ is the number of nodes and $d$ is the dimension of each feature vector. 
And $\Tilde{\mathbf{A}} = \mathbf{A} + \mathbf{I}_n$ denotes the adjacency matrix with self-loops. 
Considering a graph dataset $\mathcal{D}=\{(\mathcal{G}_1,y_1),(\mathcal{G}_2,y_2),...\}$, the target of the graph classification task is to learn a mapping function $f:\mathcal{G} \to \mathcal{Y}$, where $\mathcal{G}$ is the input graph and $\mathcal{Y}$ is corresponding label.

\textbf{Graph Neural Networks. }
In this work, we utilize Graph Convolutional Network (GCN) \cite{gcn} to construct our hierarchical pooling models.
It is obvious that our \name can also conduct the hierarchical pooling operation based on other GNNs like GIN~\cite{gin} and GAT~\cite{gat}.
A standard GCN layer can be formally formulated as:
\begin{equation}
    \mathbf{X}^{(l+1)} = \sigma (\Tilde{\mathbf{D}}^{-\frac{1}{2}} \Tilde{\mathbf{A}} \Tilde{\mathbf{D}}^{-\frac{1}{2}} \mathbf{X}^{(l)} \mathbf{W}^{(l)}),
\end{equation}
where $\sigma$ denotes the non-linear activation, $\Tilde{\mathbf{D}}=\sum_j \Tilde{\mathbf{A}}_{i,j}$ is the degree matrix of $\Tilde{\mathbf{A}}$, $\mathbf{W}^{(l)} \in \mathbb{R}^{d \times d}$ a trainable matrix in the $l$-th layer, and $\mathbf{X}^{(l+1)}$ the output of this layer and the initial feature matrix $\mathbf{X}^{(0)}=\mathbf{X}$. 

\textbf{Transformer. }
The Transformer architecture is comprised of a composition of Transformer layers.
Each layer consists of two key components: a multi-head attention module (MHA) and a position-wise feed-forward network (FFN).
The MHA module, which stacks several scaled dot-product attention layers, is focused on calculating the similarity between queries and keys. 
Specifically, let $\mathbf{H} \in \mathbb{R}^{n \times d}$ be the input of the self-attention module where $n$ is the number of input tokens and $d$ is the hidden dimension. 
The output of the self-attention module can be expressed as:
\begin{equation}
    \mathrm{Attention}(\mathbf{Q}, \mathbf{K}, \mathbf{V}) = \mathrm{softmax}\left(\frac{\mathbf{QK}^{\top}}{\sqrt{d_{out}}}\right)\mathbf{V},
    \label{eq:Attention}
\end{equation}
where $\mathbf{Q} = \mathbf{HW}_Q, \mathbf{K} = \mathbf{HW}_K, \mathbf{V} = \mathbf{HW}_V$ and $\mathbf{W}_Q \in \mathbb{R}^{d \times d}, \mathbf{W}_K \in \mathbb{R}^{d \times d}, \mathbf{W}_V \in \mathbb{R}^{d \times d}$ are trainable projection matrices.
Besides, $d_{out}$ refers to the dimension of $\mathbf{Q}$.

By concatenating multiple instances of \autoref{eq:Attention}, the multi-head attention with $m$ heads can be expressed as:
\begin{equation}
    \mathrm{MHA}(\mathbf{H}) = \mathrm{Concat}(\mathrm{head}_1, ...,\mathrm{head}_m)\mathbf{W}_O,
    \label{eq:MHSA}
\end{equation}
where $\mathrm{head}_j = \mathrm{Attention}(\mathbf{Q}_j, \mathbf{K}_j, \mathbf{V}_j)$ and $\mathbf{W}_O \in \mathbb{R}^{md \times d}$ is a parameter matrix. 
$\mathrm{Concat}(\cdot)$ represents the concatenating operation.
The output of MHA is then fed into the FFN module that consists of two linear transformations with a ReLU activation in between.  

\section{Graph Transformer Pooling}
\begin{figure}
  \includegraphics[width=8.5cm]{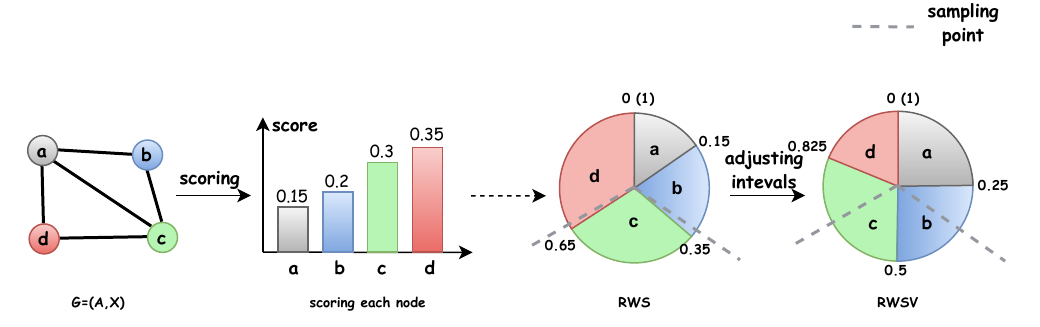}
  \caption{The example describes the sampling process of RWS and RWSV. With pooling ratio $\mu=0.5$, the sampling point is $\{\frac{1}{3}, \frac{2}{3}\}$. Obviously, RWS samples node $b$ and $d$, while RWSV samples nodes $b$ and $c$. As a comparison, the top-$K$ selection will sample nodes $c$ and $d$. We can see that RWS and RWSV can sample diversified nodes across different scoring intervals rather than only higher scoring nodes. }
  \label{fig:exp}
\end{figure}

In this section, we illustrate our proposed Graph Transformer Pooling (\name) in detail. And the architecture is depicted in \autoref{fig:frame}.

\subsection{Node Scoring}  \label{sec: Score}
Similar to existing node dropping methods, a basic step of \name is to assign a significance score to each node to measure its importance. Here we combine with the self-attention module ~\citep{transformer} to score each node from two levels: global context and local context. As the key component of the self-attention module, the attention matrix $\mathcal{A}$ can be calculated by the scaled-dot product of queries ($\mathbf{Q}$) and keys ($\mathbf{K}$): 
\begin{equation}
    \mathcal{A} = \mathrm{Softmax} \left (\frac{\mathbf{Q} \mathbf{K}^{\top}}{\sqrt{d}} \right ),
    \label{eq:sa}
\end{equation}
where  
$\mathbf{Q} \in \mathbb{R}^{n \times d}$,
$\mathbf{K} \in \mathbb{R}^{n \times d}$ 
and $d$ refers to the dimension of $\mathbf{Q}$.
Each row of $\mathcal{A}$ indicates the contribution of all input nodes to the target node. Therefore, from the perspective of global context, we compute the global significance score for all nodes as follows: 
\begin{equation}
    \mathbf{S}_g = \sigma (\mathcal{A} \mathbf{V} \mathbf{\Theta}_g ),
\end{equation}
where $\sigma$ is an activation function, $\mathbf{V} \in \mathbb{R}^{n \times d}$ is the value matrix and $\mathbf{\Theta}_g \in \mathbb{R}^{d \times 1}$ is a learnable vector. 

Besides, to measure the importance of local context, we formulate the local significance score as follows:
\begin{equation}
    \mathbf{S}_l = \sigma (\mathcal{\hat{A}} \mathbf{V} \mathbf{\Theta}_l), \\
    ~~\mathbf{\mathcal{\hat{A}}} = \mathcal{A} \odot \mathbf{\Tilde{A}}, 
\end{equation}
where $\sigma$ is an activation function and $\mathbf{\Theta}_l \in \mathbb{R}^{d \times 1}$ is a learnable vector. Here $\mathbf{S}_l$ reflects the significance of the local representations of a node's immediate neighborhood, meanwhile $\mathbf{S}_g$ captures the importance of the global representations obtained by each node interacting with all other nodes in the graph. 

Altogether, we combine these two scores to measure the importance of different nodes that explicitly take both local and global context into consideration. 
Specifically, the final scoring vector $\mathbf{S}$ is computed by taking the weighted summation of $\mathbf{S}_g$ and $\mathbf{S}_l$, which can be written as:
\begin{equation}
    \mathbf{S} = \mathrm{Softmax}( \lambda \mathbf{S}_g + (1 - \lambda) \mathbf{S}_l ),
    \label{eq:normalize}
\end{equation}
where $\lambda \in [0, 1]$ is a hyperparameter that 
trades off the global and local scores. 
For the multi-head attention module, we 
do summation after calculating the score in each head.

\subsection{Node Sampling} \label{sec: Sample}
Having acquired the final significance score of all nodes, we can select nodes via different sampling strategies. Most existing node dropping methods usually select $K$ highest scoring nodes, which may be redundant and cannot represent the original graph well. Simply performing top-$K$ selection tends to preserve nodes with similar features or topological information, meanwhile low scoring nodes that may be useful for graph representation are recklessly 
ignored. 

To gain a more representative coarsening graph, we attempt to sample nodes in a more diversified way, rather than purely high scoring nodes. Here we perform the pooling operation based on the calculated significance scores. That is to say, the probability of sampling one node is roughly equivalent to its significance score. As described in Equation \ref{eq:normalize}, we apply a $Softmax(\cdot)$ function to normalize $\mathbf{S}$, 
thus each entry $0 \leq \mathbf{S}_i \leq 1$ and $\sum_{i=0}^{n} \mathbf{S}_i = 1$. In this way, these significance scores can be regarded as probabilities. Since the corresponding nodes are discrete variables, the probability mass function can be written as:
\begin{equation}
    \mathrm{pmf} ({v}_i) = 
    \begin{cases}
        \mathbf{S}_i\ if\ i \in \{1, 2,...,n\},
        \\
        0\ \ if\  i \notin \{1, 2,...,n\},
\end{cases}
\label{eq:pmf}
\end{equation}
where $v_i$ represents the $i$-th node. 
Based on the above equation, we can further formulate the cumulative distribution function (CDF):
\begin{equation}
    \mathrm{CDF}_i = \sum_{j=1}^i \mathrm{pmf} (v_j), ~~ i\in \{1, 2, ..., n\}.
\end{equation}

Notably, due to the discrete node variable, we cannot directly take the inverse of $\mathrm{CDF}$ to perform inverse transform sampling. To sample nodes via CDF, here we introduce two ways: roulette wheel sampling ($\mathrm{RWS}$) and a variant of roulette wheel sampling ($\mathrm{RWSV}$). 
Our proposed $\mathrm{RWS}$ adopts a diversified sampling manner similar to roulette wheel selection method. For node $v_i$, its cumulative probability interval, which can be seen as the sampling interval, is $[ \mathrm{CDF}_{i-1}, \mathrm{CDF}_{i} ]$. With these $n$ intervals for all nodes, we first generate a random number $k$ from the uniform distribution $U[0,1]$ and see which interval $k$ falls in, then sample the corresponding node. The sampling process can be formulated as:  
\begin{equation}
    idx_k = \mathrm{RWS}(\mathrm{CDF}, k),
\end{equation}
where $idx_k$ is the node index selected by $k$. To guarantee the deterministic of training and inference, we leverage a fixed sampling scheme by choosing $k$ as $\{\frac{1}{M + 1}, \frac{2}{M + 1},..., \frac{M}{M + 1} \}$, among which $M=\lceil \mu * n \rceil$ is the number of nodes to sample and $\mu$ is the pooling ratio.

By considering the significance score as probability and combining it with a regular sampling scheme, we find $\mathrm{RWS}$ can efficiently preserve low-scoring nodes in most instances. But in practice, low-scoring nodes will be overlooked in some cases, 
especially when the number of such low-scoring nodes is rare.

To further enhance the preference for low-scoring nodes, we present another sampling method based on $\mathrm{CDF}$ which can be actually interpreted as a variant of $\mathrm{RWS}$. In this approach, with a given $k$, we directly take the index of the node with the nearest cumulative score as the sampling index. Named it as $\mathrm{RWSV}$, the sampling procedure can be shown as:
\begin{equation}
    idx_k = \mathrm{RWSV} (\mathrm{CDF}, k), 
\end{equation}
where $\mathrm{RWSV}(\cdot)$ represents the sampling operation that gets the target node whose cumulative score has a minimum gap with $k$. Same as $\mathrm{RWS}$, here we also leverage the fixed sampling scheme on the value of $k$. As for why $\mathrm{RWSV}$ can be regarded as a variant of $\mathrm{RWS}$, the reason is that $\mathrm{RWSV}$ only slightly adjusts the sampling interval of nodes. As described above, the sampling interval of $v_i$ in $\mathrm{RWS}$ is $[ \mathrm{CDF}_{i-1}, \mathrm{CDF}_{i} ]$. While for $\mathrm{RWSV}$, we find the sampling interval for $v_i$ is partitioned as: 
\begin{equation} 
    {interval} ({v}_i) = 
    \begin{cases}
        [0, \frac{\mathrm{CDF}_i + \mathrm{CDF}_{i+1}}{2}]\ \ if\ \ i = 0,
        \\
        [\frac{\mathrm{CDF}_{i-1} + \mathrm{CDF}_i}{2}, \frac{\mathrm{CDF}_i + \mathrm{CDF} _{i+1}}{2}]\ \ if\ \ i \in \{2,...,n-1\},
        \\
        [\frac{\mathrm{CDF}_{i-1} + \mathrm{CDF}_i}{2}, 1]\ \ if\  i = n.
\end{cases}
    \label{eq:pmf}
\end{equation}

Compared with $\mathrm{RWS}$, the sampling intervals for low-scoring nodes are clearly enlarged. Thus $\mathrm{RWSV}$ has the ability to preserve more low-scoring nodes. 
Figure \ref{fig:exp} provides an example about the sampling process of RWS and RWSV.
Practically, we find that $\mathrm{RWSV}$ outperforms $\mathrm{RWS}$ on most chosen benchmarks, which also verifies the effectiveness of the adjusted sampling intervals. 

During the sampling process, another problem is that a node may be sampled more than once.
Therefore, we concretely provide a re-sampling strategy to ensure the uniqueness of the pooled nodes. The basic re-sampling rule is that for both $\mathrm{RWS}$ and $\mathrm{RWSV}$, if the target node has been selected before, we utilize the nearest left neighboring node on the roulette wheel for replacement. 
After the sampling and re-sampling operation, we can obtain $M$ unique nodes, and their indices can be expressed as $\mathbf{idx} = \{idx_1, idx_2, ..., idx_M\}$.

\begin{figure}
  \includegraphics[width=8.5cm]{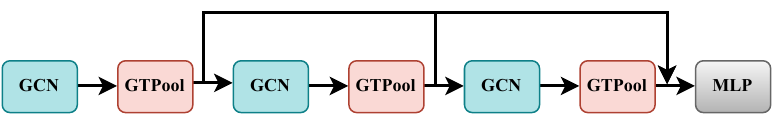}
  \caption{The hierarchical pooling architecture of \name. Following the design of previous works, the architecture is constructed by three GCN layers and each layer is followed with a \name layer. }
  \label{fig:hp}
\end{figure}

\subsection{Graph Transformer Pooling}
Based on the proposed node scoring and node sampling method, we continue to introduce the pooling process within the Transformer pooling layer to get a coarsened graph $\mathcal{G}^{'}=(\mathbf{A}^{'}, \mathbf{X}^{'})$. 

\textbf{Update the Adjacency Matrix. } 
With the indices of the pooled nodes, we can update its adjacency matrix by selecting the corresponding rows and columns. The new adjacency matrix $\mathbf{{A}}^{'}$ can be expressed as:
\begin{equation}
    \mathbf{{A}}^{'} = \mathbf{{A}}_{\mathbf{idx}, \mathbf{idx}}, 
\end{equation}
where $\mathbf{{A}}^{'} \in \mathbb{R}^{M \times M}$. 

\textbf{Update the Attribute Matrix. }
Here, we first utilize the obtained indices to generate a refined attention matrix by preserving the rows of the corresponding indices, which can be formulated as:  
\begin{equation}
    \mathcal{A}^{'} = \mathcal{A}_{\mathbf{idx},:},
\end{equation}
where $\mathcal{A}^{'} \in \mathbb{R}^{M \times n}$ denotes the refined attention matrix. Then, to acquire $M$ pooled nodes, we leverage $\mathcal{A}^{'}$ to replace the original attention matrix $\mathcal{A}$ such that:
\begin{equation}
    \mathbf{\hat{X}} = \mathcal{A}^{'} \mathbf{V} + \mathbf{X}_{\mathbf{idx}, :},
    \label{eq:MHA}
\end{equation}
where $\mathbf{\hat{X}}, \mathbf{X}_{\mathbf{idx}, :} \in \mathbb{R}^{M \times d}$ and $\mathbf{X}_{\mathbf{idx}, :}$ is the residual bias term that 
denotes the original attribute of the pooled nodes. Notably, by Equation \ref{eq:MHA}, each sampled node can sincerely react with all other nodes on the original graph, thus effectively capturing long-range dependencies. After that, the output $\mathbf{\hat{X}}$ is fed into a standard FFN module in the Transformer layer which consists of two linear layers and a GeLU activation. The process can be formally described as:
\begin{equation}
    \mathbf{X}^{'} = \mathrm{FFN}(\mathrm{LN}(\mathbf{\hat{X}})) + \mathbf{\hat{X}},
\end{equation}
where $\mathrm{LN}(\cdot)$ indicates layer normalization and $\mathbf{X}^{'} \in \mathbb{R}^{M \times d}$ is the final output embeddings of the sampled nodes.

In conclusion, with the two updating steps above, we can explicitly generate a new refined graph $\mathcal{G}^{'}=(\mathbf{A}^{'}, \mathbf{X}^{'})$. 

\subsection{Model Architecture}
In this subsection, we concretely present the pooling architecture that we utilize in this paper for graph classification.
Based on the proposed sampling method RWS and its variant RWSV, we name our pooling operators as \name and \vname, respectively.
Here, we take \name as the example.

As shown in Figure \ref{fig:hp},  following the setting of previous works~\cite{sagpool, gmt, mincutpool}, 
there are three blocks in the architecture, each of which consists of a graph convolutional layer and a \name layer.
For the graph-level representation, we first readout the output of each \name layer, and then get their summation, which can be formulated as:
\begin{equation}
    \mathbf{h}_g = \sum_{i=1}^l Readout(GTPool_i(GCN_i(\mathbf{X}_i, \mathbf{A}_i)),
\end{equation}
where $\mathbf{h}_g$ is the graph representation, $l$ is 3 in Figure \ref{fig:hp}, $\mathbf{X}_i$ and $\mathbf{A}_i$ are the input feature matrix and adjacency matrix to the $i$-th layer, respectively.

With this hierarchical architecture, we can obtain more expressive graph representations than existing node dropping methods due to the following merits: (1) the RWS (or RWSV) can preserve diverse representative nodes rather than $k$ higher scoring nodes with similar attributes; (2) the pooled nodes have captured both local and global information while existing methods only aggregate local information.


\section{Experiments}
\begin{table*}
  \caption{Test performance on 11 benchmarks from TUDatasets and OGB. The shown results are mean ($\%$) and deviation ($\%$) over 10 different runs.
  The best results appear in bold and hyphen (-) indicates out-of-resources that take more than 10 days.
  }
  \label{tab:tud_res}
  \setlength\tabcolsep{1.0mm}{
	\scalebox{0.68}{
  \begin{tabular}{lccccccccccc}
    \toprule
    Model & MUTAG $\uparrow$ & ENZYMES $\uparrow$ & PROTEINS $\uparrow$ & PTC-MR $\uparrow$ & Synthie $\uparrow$ & IMDB-B $\uparrow$ & IMDB-M $\uparrow$ & HIV $\uparrow$ & TOX21 $\uparrow$ & TOXCAST $\uparrow$ & BBBP $\uparrow$\\
    \midrule
    GCN & 69.50 $\pm$ 1.78 & 43.52 $\pm$ 1.12 & 73.24 $\pm$ 0.73 & 55.72 $\pm$ 1.64 & 58.61 $\pm$ 1.62 & 73.26 $\pm$ 0.46 & 50.39 $\pm$ 0.41 & 76.81 $\pm$ 1.01 & 75.04 $\pm$ 0.80 & 60.63 $\pm$ 0.51 & 65.47 $\pm$ 1.83 \\
    GIN & 81.39 $\pm$ 1.53 & 40.19 $\pm$ 0.84 & 71.46 $\pm$ 1.66 & 56.39 $\pm$ 1.25 & 59.20 $\pm$ 0.73 & 72.78 $\pm$ 0.86 & 48.13 $\pm$ 1.36 & 75.95 $\pm$ 1.35 & 73.27 $\pm$ 0.84 & 60.83 $\pm$ 0.46 & 67.65 $\pm$ 3.00 \\
    \midrule
    Set2Set & 69.89 $\pm$ 1.94 & 42.92 $\pm$ 2.05 & 73.27 $\pm$ 0.85 & 54.52 $\pm$ 1.69 & 46.12 $\pm$ 2.71 & 73.10 $\pm$ 0.48 & 50.15 $\pm$ 0.58 &  73.42 $\pm$ 2.34 & 73.42 $\pm$ 0.67 & 59.76 $\pm$ 0.65 & 64.43 $\pm$ 2.16 \\
    SortPool & 71.94 $\pm$ 3.55 & 36.17 $\pm$ 2.58 & 73.17 $\pm$ 0.88 & 52.62 $\pm$ 2.11 & 50.18 $\pm$ 1.77 &72.12 $\pm$ 1.12 & 48.18 $\pm$ 0.63 & 71.82 $\pm$ 1.63 & 69.54 $\pm$ 0.75 & 58.69 $\pm$ 1.71 & 65.98 $\pm$ 1.70 \\
    DiffPool & 79.22 $\pm$ 1.02 & 51.27 $\pm$ 2.89 & 73.03 $\pm$ 1.00 & 55.26 $\pm$ 3.84 & 62.75 $\pm$ 0.74 & 73.14 $\pm$ 0.70 & 51.31 $\pm$ 0.72 & 75.64 $\pm$ 1.86 & 74.88 $\pm$ 0.81 & 62.28 $\pm$ 0.56 & 68.25 $\pm$ 0.96 \\
    $\mathrm{SAGPool}_{g}$ & 76.78 $\pm$ 2.12 & 36.30 $\pm$ 2.51 & 72.02 $\pm$ 1.08 & 54.38 $\pm$ 1.96 & 51.17 $\pm$ 1.71 & 72.16 $\pm$ 0.88 & 49.47 $\pm$ 0.56 & 74.56 $\pm$ 1.69 & 71.10 $\pm$ 1.06 & 59.88 $\pm$ 0.79 & 65.16 $\pm$ 1.93 \\
    $\mathrm{SAGPool}_{h}$ & 73.67 $\pm$ 4.28 &  34.12 $\pm$ 1.75 & 71.56 $\pm$ 1.49 & 53.82 $\pm$ 2.44 & 51.75 $\pm$ 0.66 & 72.55 $\pm$ 1.28 & 50.23 $\pm$ 0.44 & 71.44 $\pm$ 1.67 & 69.81 $\pm$ 1.75 & 58.91 $\pm$ 0.80 & 63.94 $\pm$ 2.59 \\
    TopKPool & 67.61 $\pm$ 3.36 & 29.77 $\pm$ 2.74 & 70.48 $\pm$ 1.01 & 55.59 $\pm$ 2.43 & 50.64 $\pm$ 0.47 & 71.74 $\pm$ 0.95 & 48.59 $\pm$ 0.72 & 72.27 $\pm$ 0.91 & 69.39 $\pm$ 2.02 & 58.42 $\pm$ 0.91 & 65.19 $\pm$ 2.30 \\
    MincutPool & 79.17 $\pm$ 1.64 & 25.33 $\pm$ 1.47 & 74.72 $\pm$ 0.48 & 54.68 $\pm$ 2.45 & 52.61 $\pm$ 1.44 & 72.65 $\pm$ 0.75 & 51.04 $\pm$ 0.70 & 75.37 $\pm$ 2.05 & 75.11 $\pm$ 0.69 & 62.48 $\pm$ 1.33 & 65.97 $\pm$ 1.13 \\
    StructPool & 79.50 $\pm$ 1.75 & 55.60 $\pm$ 1.94 & 75.16 $\pm$ 0.86 & 55.72 $\pm$ 1.41 & 55.67 $\pm$ 0.92 & 72.06 $\pm$ 0.64 & 50.23 $\pm$ 0.53  & 75.85 $\pm$ 1.81 & 75.43 $\pm$ 0.79 & 62.17 $\pm$ 0.61 & 67.01 $\pm$ 2.65 \\
    ASAP & 77.83 $\pm$ 1.49 & 20.10 $\pm$ 1.13 & 73.92 $\pm$ 0.63 & 55.68 $\pm$ 1.45 & 47.93 $\pm$ 1.36 & 72.81 $\pm$ 0.50 & 50.78 $\pm$ 0.75 & 72.86 $\pm$ 1.40 & 72.24 $\pm$ 1.66 & 58.09 $\pm$ 1.62 & 63.50 $\pm$ 2.47 \\
    HaarPool & 66.11 $\pm$ 1.50 & - & - & - & - & 73.29 $\pm$ 0.34 & 49.98 $\pm$ 0.57 & - & - & - & 66.11 $\pm$ 0.82 \\
    GMT & 83.44 $\pm$ 1.33 & 37.38 $\pm$ 1.52 & 75.09 $\pm$ 0.59 & 55.41 $\pm$ 1.30 & 52.65 $\pm$ 0.09 & 73.48 $\pm$ 0.76 & 50.66 $\pm$ 0.82 & 77.56 $\pm$ 1.25 & 77.30 $\pm$ 0.59 & \bf{65.44} \textbf{$\pm$} \bf{0.58} & 68.31 $\pm$ 1.62 \\
    \midrule
    \textbf{\name} & 87.64 $\pm$ 0.34 & 61.09 $\pm$ 0.85 & 75.42 $\pm$ 0.09 & \bf{61.17} \textbf{$\pm$} \bf{0.18} & 67.50 $\pm$ 0.61 & 74.04 $\pm$ 0.33 & 50.89 $\pm$ 0.22 & 77.94 $\pm$ 0.46 & 77.32 $\pm$ 0.12 & 65.34 $\pm$ 0.87 & 70.15 $\pm$ 0.29 \\
    \textbf{\vname} & \bf{88.15} \textbf{$\pm$} \bf{0.45} & \bf{61.31} \textbf{$\pm$} \bf{0.81} & \bf{75.68} \textbf{$\pm$} \bf{0.06} & 60.91 $\pm$ 0.09 & \bf{68.08} \textbf{$\pm$} \bf{0.75} & \bf{74.23} \textbf{$\pm$} \bf{0.44} & \bf{51.49} \textbf{$\pm$} \bf{0.18} & \bf{78.36} \textbf{$\pm$} \bf{0.35} & \bf{77.45} \textbf{$\pm$} \bf{0.56} & {65.18} {$\pm$} {1.12} & \bf{70.68} \textbf{$\pm$} \bf{0.33}\\
  \bottomrule
    \end{tabular}
}
}
\end{table*}

In this section, we evaluate our proposed methods: \name (with RWS) and \vname (with RWSV) on several graph benchmarks for graph classification. 
Various strong baselines from the field of graph pooling are elaborately selected to compare with our methods. 
To further identify the factors that drive the performance, we conduct ablation studies to analyze the contribution of each key component in our \name.

\subsection{Experimental Setup}
We first briefly introduce the datasets, baselines, and basic configurations used in our experiments.

\textbf{Datasets.} 
We conduct experiments on 11 widely used datasets to validate the effectiveness of our model.  
Specifically, we adopt seven datasets from TUDatasets~\cite{tudataset}, including MUTAG, ENZYMES, PROTEINS, PTC-MR, Synthie, IMDB-B and IMDB-M with accuracy for evaluation metric. Besides, we employ four molecule datasets from Open Graph Benchmark~\cite{ogb}, including HIV, Tox21, ToxCast and BBBP with ROC-AUC for evaluation metric.    
The detailed statistics are summarized in Appendix \ref{ds_stat}.

\textbf{Baselines.} 
We first consider two popular backbones for comparison: GCN~\cite{gcn} and GIN~\cite{gin}. 
Then, we utilize four node dropping methods that drop nodes with lower scores, including SortPool~\cite{sortpool}, SAGPool~\cite{sagpool}, TopKPool~\cite{topkpool} and ASAP~\cite{asap}. 
As another mainstream graph pooling method, we also select four node clustering models that group a set of nodes into a cluster, including DiffPool~\cite{diffpool}, MinCutPool~\cite{mincutpool}, HaarPool~\cite{haarpool} and StructPool~\cite{structpool}. Besides, we additionally choose two commonly used baselines: Set2Set~\cite{set2set} and GMT~\cite{gmt}. 
Set2Set utilizes a recurrent neural network to encode all nodes with content-based attention. GMT formulates graph pooling as a multiset encoding problem and can be actually regarded as a combination of node clustering method with Transformer. 

\textbf{Configurations.} 
For the chosen benchmarks from TUDatasets, we follow the dataset split in \cite{sortpool, gin, gmt} and conduct 10-fold cross-validation to evaluate the model performance.  
The initial node attribute we use is in line with the fair comparison setup in \cite{fair}. 
For datasets from OGB, we follow the original feature extraction and the provided standard dataset splitting in \cite{ogb}. More experimental and hyperparameter details are described in Appendix \ref{exp}.

\subsection{Performance Comparison}
We compare the performance of our proposed \name and \vname with the baseline methods on 11 benchmark datasets for the graph classification task.

For the seven benchmarks from TUDatasets, we report the average accuracy with standard deviation after running 10-fold cross-validation ten different times. 
The experimental results are summarized in Table \ref{tab:tud_res}, and the best results appear in bold. 
First, it is evident that both \name and \vname obtain better or competitive performance as compared to these popular pooling methods. 
To be specific, our proposed methods achieve approximate $4.71\%, 5.71\%, 4.78\%, 5.32\%$ higher accuracy over the best baselines on MUTAG, ENZYMES, PTC-MR and Synthie datasets respectively, demonstrating the effectiveness and superiority of our methods.
Next, one can see that the proposed models consistently outperform DiffPool and MincutPool on all the datasets, which reveals the importance of combining local context and global context.
Compared with SAGPool, TopKPool, and ASAP, the improved performance can be attributed to the better capacity of our methods to preserve more representative nodes. 
Especially, the performance of our methods significantly surpasses GMT, indicating that hierarchical pooling brings more improvement than global pooling when cooperating with Transformer.
Besides, we can also observe that GMT, \name and \vname gain better performance than other graph pooling counterparts, showing the necessity of introducing Transformer to the graph pooling domain. 

To validate the effectiveness of our proposed method on diverse datasets, we further conduct experiments on four molecule datasets from the Open Graph Benchmark.
By running each experiment 10 times and taking the average and standard deviation of the corresponding metric, the obtained experimental results are reported in Table \ref{tab:tud_res}. 
Overall, our proposed \name and \vname achieve higher performance than other baseline models. 
It is worth emphasizing that when compared to other node dropping pooling methods and GMT, which is a Transformer-based global pooling method, the superior performance of our \name steadfastly confirms the effectiveness of combining Transformer with node dropping pooling to hierarchically learn graph representations. 

\subsection{Effectiveness of RWS \& RWSV }
Due to the plug-and-play property of our novel sampling schemes (RWS and RWSV), 
we further conduct extensive experiments to verify the efficacy by applying them to three typical node dropping methods,  including $\mathrm{SAGPool}_h$, TopKPool and ASAP. 
We choose PROTEINS, IMDB-BINARY and BBBP as the benchmarks, and the experimental results are shown in Table \ref{tab:rws_com}.

By simply replacing the top-$K$ module with our RWS or RWSV, we can observe that these three modified models achieve significant enhancement on the selected datasets. 
Specifically, the average improvements across the selected datasets over three node dropping pooling methods are approximately $1.73\%$ on $\mathrm{SAGPool}_h$, $1.41\%$ on TopKPool and $0.84\%$ on ASAP. 
These improvements strongly demonstrate that both RWS and RWSV are more capable of preserving representative nodes than the simply top-$K$ selection. 
In addition, we can also see that RWSV gains better performance in most cases apart from ASAP on PROTEINS and IMDB-BINARY.
Note that RWSV still obtains competitive performance on these two datasets.
The empirical results clearly show the importance of adjusting the sampling intervals. 
\begin{table}
  \caption{Test performance on PROTEINS, IMDB-BINARY and BBBP when applying RWS and RWSV to 
  \textbf{SAGPool$_h$}, TopKPool and ASAP. The shown results are mean ($\%$) and deviation ($\%$) over 10 different runs.
  The best results appear in bold. 
  }
  \label{tab:rws_com}
  \setlength\tabcolsep{1.4mm}{
	\scalebox{0.7}{
  \begin{tabular}{llccc}
    \toprule
    Model & Method & PROTEINS & IMDB-B & BBBP\\
    \midrule
    $\mathrm{SAGPool}_h$ & top-$K$ & 71.56 $\pm$ 1.49 & 72.55 $\pm$ 1.28 & 63.94 $\pm$ 2.59 \\
    $\mathrm{SAGPool}_h$ & RWS & 72.07 $\pm$ 2.46 & 73.01 $\pm$ 0.69 & 65.76 $\pm$ 0.52 \\
    $\mathrm{SAGPool}_h$ & RWSV & \bf{73.51} \textbf{$\pm$} \bf{1.06} & \bf{73.22} \textbf{$\pm$} \bf{0.36} & \bf{66.51} \textbf{$\pm$} \bf{1.29} \\
    \midrule
    TopKPool & top-$K$ & 70.48 $\pm$ 1.01 & 71.74 $\pm$ 0.95 & 65.19 $\pm$ 2.30 \\
    TopKPool & RWS & 71.42 $\pm$ 1.55 & 72.66 $\pm$ 0.81 & 65.40 $\pm$ 0.87 \\
    TopKPool & RWSV & \bf{72.04} \textbf{$\pm$} \bf{1.44} & \bf{73.35} \textbf{$\pm$} \bf{0.67} & \bf{66.24} \textbf{$\pm$} \bf{1.75} \\
    \midrule
    ASAP & top-$K$ & 73.92 $\pm$ 0.63 & 72.81 $\pm$ 0.50 & 63.50 $\pm$ 2.47 \\
    ASAP & RWS & \bf{74.17} \textbf{$\pm$} \bf{1.12} & \bf{73.79} \textbf{$\pm$} 0.71 & 64.22 $\pm$ 1.65 \\
    ASAP & RWSV & 74.06 $\pm$ 0.68 & 73.60 $\pm$ 0.70 & \bf{64.78} \textbf{$\pm$} \bf{1.04} \\
    \bottomrule
    \end{tabular}
}
}
\end{table}

\subsection{Ablation Studies}
\begin{table}
  \caption{Ablation studies of \vname on the PROTEINS, IMDB-BINARY and BBBP datasets for graph classification. 
  The best results appear in bold. 
  }
  \label{tab:abl_res}
  \setlength\tabcolsep{1.0mm}{
	\scalebox{0.7}{
  \begin{tabular}{lccc}
    \toprule
    Model & PROTEINS & IMDB-B & BBBP\\
    \midrule
    \vname & \bf{75.68} \textbf{$\pm$} \bf{0.06} & \bf{74.23} \textbf{$\pm$} \bf{0.44} & \bf{70.68} \textbf{$\pm$} \bf{0.33} \\
    \midrule
    w/o local significance score & 73.03 $\pm$ 0.70 & 73.38 $\pm$ 0.28 & 69.47 $\pm$ 0.56\\
    w/o global significance score & 75.40 $\pm$ 0.27 & 73.58 $\pm$ 0.29 & 70.33 $\pm$ 0.44\\
    \midrule
    top-$K$ selection & 73.70 $\pm$ 1.44 & 73.20 $\pm$ 0.42 & 67.15 $\pm$ 0.69\\
  \bottomrule
    \end{tabular}
}
}
\end{table}
In this subsection, we perform a series of ablation studies to verify the key component that drives high performance in our proposed \vname. 
The ablation results are reported in Table \ref{tab:abl_res}.

\textbf{Significance Score.}
In order to investigate the effectiveness of the global and local significance scores, we compare our proposed \vname to its variants without global and local significance score modules, respectively.
The experimental results show that reasonably incorporating both global and local significance scores can significantly improve the performance of \vname, proving the necessity of considering both. 
Apparently, one can observe that the performance degradation without the global significance score is more severe than without the local score. 
In some way, this phenomenon signifies the importance of the local context outweighs the global context.

\textbf{Roulette Wheel Sampling.}
To explore whether RWSV brings distinctive enhancement, we conduct a comparative experiment between the proposed RWSV and the most commonly used top-$K$ selection.
As described before, top-$K$ selection only samples the higher scoring nodes, discarding all lower scoring nodes.
From Table \ref{tab:abl_res}, we can observe that RWSV outperforms top-$K$ selection by a large margin, showing the superiority of sampling nodes in an adaptive way.

\subsection{Parameter Analysis}
\begin{figure}
  \includegraphics[width=7cm]{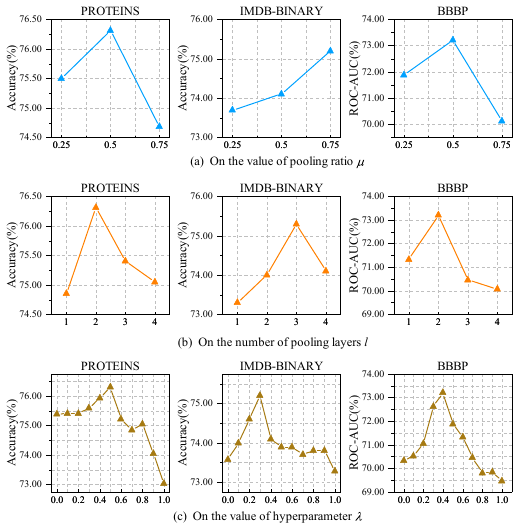}
  \caption{Parameter studies on the pooling ratio $\mu$, the number of pooling layers $l$ and the weight $\lambda$.}
  \label{fig:pa}
\end{figure}
We further conduct a variety of studies to investigate the sensitivity of our proposed \vname on three key parameters: the pooling ratio $\mu$, the number of pooling layers $l$ and the weight $\lambda$ that decides which score matters more. 
The empirical results on PROTEINS, IMDB-BINARY and BBBP by setting different values are illustrated as follows.

\textbf{On the value of Pooling Ratio $\mu$. }
Since the value of pooling ratio $\mu$ is closely related to the performance on different datasets, we first test the effect on the value of pooling ratio $\mu$ selecting from $\{0.25, 0.5, 0.75\}$ and report the performance of \vname. 
The results are shown in Figure \ref{fig:pa} (a). 
Notably, we can observe that \vname achieves better performance on both PROTEINS and BBBP when setting $\mu$ as $0.5$, and the performance severely drops about $2\%-3\%$ when $\mu=0.75$.
The reason can be attributed to the fact that preserving too many nodes will inevitably introduce additional redundancies. 
Besides, with the increment of $\mu$, the performance on IMDB-BINARY continues to rise because the average number of nodes is lower than PROTEINS and BBBP. Thus, a larger $\mu$ brings more affluent information. 

\textbf{On the number of pooling layers $l$. }
We further conduct experiments to explore the performance of \name with different pooling times. 
With all other parameters fixed, we vary the number of pooling layers $l$ from $1$ to $4$ and report the results, as shown in Figure \ref{fig:pa} (b).
In general, the performance curves on three datasets present a moderate increase trend at first, then start to decline.
The results demonstrate that it is essential to pick up a suitable pooling time, since stacking too many Transformer-based \vname layers may easily lead to over-fitting. 

\textbf{On the value of weight $\lambda$. }
A larger $\lambda$ signifies concentrating more on the global context and paying less attention to the local context. 
To find out how $\lambda$ actually influences the model performance, we conduct a series of studies by varying $\lambda$ from $0$ to $1$ with a step length as 0.1.
The experimental results are shown in Figure \ref{fig:pa} (c).
We interpret the results from two perspectives. 
First, considering global context can practically bring performance gain for all three datasets. This verifies the effectiveness of the global significance score. 
Second, with the increment of $\lambda$, the performance on three datasets increases first and then starts to degrade.
All in all, these datasets achieve the best performance when $\lambda$ is from 0.3 to 0.5, which demonstrates that properly considering the global context is beneficial for the model performance.

\subsection{Computational Efficiency}
\begin{table}
  \caption{Speedup of \vname. For \vname models trained on PROTEINS, IMDB-BINARY and BBBP, we find that the combination with Transformer increases minimal overhead. Speedup refers to the training iteration speed comparing with SAGPool and a larger number signifies a faster training speed. 
  }
  \label{tab:speedup}
  \setlength\tabcolsep{1.6mm}{
	\scalebox{0.7}{
  \begin{tabular}{lcccc}
    \toprule
    Dataset & Model & Forward time (ms) & Backward time (ms) & Speedup \\
    \midrule
     & $\mathrm{SAGPool}_h$ & 26.62 $\pm$ 2.25 & 31.82 $\pm$ 3.02 & 1 $\times$ \\ 
     & TopKPool & 21.84 $\pm$ 1.60 & 24.81 $\pm$ 1.72 & 1.25 $\times$\\
     PROTEINS & ASAP & 188.65 $\pm$ 4.75 & 114.71 $\pm$ 5.12 & 0.19 $\times$ \\
     & GMT & 105.46 $\pm$ 6.52 & 126.44 $\pm$ 5.35 & 0.25 $\times$\\
     & \vname & 24.11 $\pm$ 3.22 & 20.54 $\pm$ 2.80 & 1.31 $\times$\\

     \midrule
     & $\mathrm{SAGPool}_h$ & 20.85 $\pm$ 3.73 & 20.95 $\pm$ 1.44 & 1 $\times$ \\ 
     & TopKPool & 26.32 $\pm$ 2.07 & 25.84 $\pm$ 1.65 & 0.80 $\times$\\
     IMDB-B & ASAP & 152.18 $\pm$ 5.37 & 68.08 $\pm$ 3.40 & 0.19 $\times$\\
     & GMT & 66.32 $\pm$ 3.31 & 82.61 $\pm$ 3.77 & 0.28 $\times$\\
     & \vname & 28.37 $\pm$ 1.95 & 16.56 $\pm$ 4.11 & 0.93 $\times$\\

     \midrule
     & $\mathrm{SAGPool}_h$ & 13.94 $\pm$ 4.12 & 16.86 $\pm$ 3.09 & 1 $\times$ \\ 
     & TopKPool & 9.66 $\pm$ 3.12 & 14.36 $\pm$ 2.48 & 1.28 $\times$\\
     BBBP & ASAP & 130.78 $\pm$ 6.43 & 78.22 $\pm$ 4.55 & 0.15 $\times$\\
     & GMT & 43.71 $\pm$ 2.44 & 158.24 $\pm$ 5.89 & 0.15 $\times$\\
     & \vname & 23.84 $\pm$ 3.58 & 14.98 $\pm$ 2.78 & 0.79 $\times$\\
  \bottomrule
    \end{tabular}
}
}
\end{table}
To quantitatively evaluate the overhead that our \vname brings over typically hierarchical node dropping pooling methods, We report the forward pass time and backward pass time per iteration. 
The statistical results are shown in Table \ref{tab:speedup}. 
In order to ensure the fairness of the comparison, we first normalize these models to have approximately similar parameter numbers.
Evidently, we can observe that \vname is actually slightly faster to train on the PROTEINS dataset than $\mathrm{SAGPool}_h$ and TopKPool, meanwhile exceeding ASAP and GMT by a large margin.
For the IMDB-BINARY and BBBP datasets, the training time of \vname is  
$7 - 21\%$ slower than $\mathrm{SAGPool}_h$ and greatly surpasses ASAP and GMT with more than $64\%$. 

\subsection{Scalability}
\begin{table}
  \caption{Scalability of \vname to large graphs. We profile our \vname on these randomly generated graphs and report the runtime in milliseconds. 
  }
  \label{tab:scalability}
  \setlength\tabcolsep{2.2mm}{
	\scalebox{0.7}{
  \begin{tabular}{llcccc}
    \toprule
    & &  \multicolumn{4}{c}{Edge Density} \\ 
    Node Count & Model & 20\% & 40\% & 60\% & 80\%\\
    \midrule
    \multirow{4}{1cm}{500} & $\mathrm{SAGPool}_h$ & \bf{19.63} & \bf{23.29} & \bf{28.50} & 34.50\\
    & ASAP & 217.51 & 249.49 & OOM & OOM\\
    &  GMT & 25.27 & 29.02 & 34.12 & 37.66 \\
    & \vname & 30.31 & 31.74 & 33.22 & \bf{34.21} \\
    
    \midrule
    \multirow{4}{1cm}{1000} & $\mathrm{SAGPool}_h$ & \bf{34.49} & 53.48 & 73.26 & 92.76 \\
    & ASAP & OOM & OOM & OOM & OOM\\
    &  GMT & 42.30 & 63.72 & 90.87 & 115.31 \\
    & \vname & 41.92 & \bf{47.07} & \bf{52.16} & \bf{57.08} \\
    
    \midrule
    \multirow{4}{1cm}{1200} & $\mathrm{SAGPool}_h$ & 48.23 & 71.54 & OOM & OOM \\
    & ASAP & OOM & OOM & OOM & OOM\\
    &  GMT & 49.47 & 86.18 & 124.05 & OOM \\
    & \vname & \bf{48.03} & \bf{57.10} & \bf{63.57} & OOM \\
  \bottomrule
    \end{tabular}
}
}
\end{table}
To investigate whether our \vname can scale to large graphs over 100 nodes, we report the runtime of our model on different microbenchmarks with varying node counts and edge densities.
These microbenchmarks are comprised of randomly generated Erdos-Renyi graphs with a varying number of nodes from $\{500, 1000, 1200\}$ and edge density from $\{20\%, 40\%, 60\%, 80\%\}$.
We elaborately choose $\mathrm{SAGPool}_h$, ASAP and GMT as baselines and the summarized results are shown in Table \ref{tab:scalability}.
We can observe that as the number of nodes and edge density increase, our \vname can scale to larger graphs with less runtime.
Notably, both $\mathrm{SAGPool}_h$ and ASAP meet out of memory errors (OOM) when the number of nodes is 1200 and edge density is $80\%$, but \vname can effectively handle these graphs within roughly a half runtime than GMT.

\section{Conclusion}
In this paper, we proposed \name, a hierarchical and Transformer-based node dropping pooling method for graph structured data.
Our method has the following features: hierarchical pooling, consideration of both local and global context, adaptive sampling and be able to capture long-range dependencies for each pooled node.
A \name layer is actually constructed based on a Transformer encoder layer combining with a scoring module and a sampling module.
Specifically, \name first leverages the global and local context to score each node, then uses an adaptive roulette wheel sampling (RWS) module to diversely select nodes rather than samping only $M$ higher scoring nodes.
With the indices of the sampled nodes, we can refine the attention matrix from $n \times n$ to $M \times n$ and feed it into the subsequent modules of the Transformer.
In this way, by iteratively stacking GNN and GTPool layers, the pooled nodes can effectively capture both long-range and local relationships during the pooling process.
We then conduct extensive experiments on 11 benchmark datasets and find that \name leads to impressive improvements upon multiple the state of the art pooling methods.

The main limitation of \name is the same as the original Transformer that it suffers from the quadratic complexity of the self-attention mechanism, making it hard to apply to large graphs.
In future work, we will attempt to combine efficient Transformers with node pooling methods.


\bibliographystyle{ACM-Reference-Format}
\bibliography{07_Ref}

\newpage
\appendix

\section{Datasets and Baselines}

\subsection{Datasets} \label{ds_stat}
The statistics of the datasets we adopt in experiments are shown in \autoref{tab:stat}. There are 11 datasets, with number of graph classes ranging from 2 to 617. 

\subsection{Baselines}
This subsection concretely introduces the baseline methods utilized in comparison for the graph classification task.

\begin{itemize}

    \item \textbf{GCN}~\cite{gcn} first introduces a hierarchical propagation process to aggregate information from $1$-hop neighborhood layer by layer and we take it as the mean pooling baseline in this paper.

    \item \textbf{GIN}~\cite{gin} is provably as powerful as the Weisfeiler-Lehman (WL) graph isomorphism test and is employed as the sum pooling baseline for comparison.

    \item \textbf{Set2Set}~\cite{set2set} is a set pooling baseline that utilizes a recurrent neural network to encode all nodes with content-based attention.

    \item \textbf{SortPool}~\cite{sortpool} is a node dropping baseline that  directly drops nodes by sorting their representations generated from the previous GNN layers.

    \item \textbf{SAGPool}~\cite{sagpool} introduces the self-attention mechanism to learn the node-wise importance scores. Then, the top-$K$ based node sampling strategy is adopted for selecting the important nodes for graph pooling.

    \item \textbf{TopKPool}~\cite{topkpool} expands upon the U-Net framework and leverages the graph convolutional layer and skip-connection to maintain high-level information of the input graph, which is further used for graph pooling. 

    \item \textbf{MincutPool}~\cite{mincutpool} leverages spectral clustering techniques to identify clusters or subgraphs within the original graph, and then generates clusters by spectral clustering which are treated as nodes in a new graph, achieving the graph pooling operation.

    \item \textbf{StructPool}~\cite{structpool} utilizes Conditional Random Fields to perform structured graph reduction. By considering the dependencies between nodes, it creates a more informative and structured representation of the original graph.

    \item \textbf{ASAP}~\cite{asap} adopts the hierarchical architecture and utilizes the combination of local and global information for calculating the important scores of nodes.  

    \item \textbf{HaarPool}~\cite{haarpool} is a node clustering based method which computes following a chain of sequential clusterings of the input graph. The compressive Haar transformation is adopted for pooling nodes within the same cluster in the frequency domain. 

    \item \textbf{GMT}~\cite{gmt} treats a graph pooling problem as a multiset encoding problem and leverages the self-attention mechanism to learn the representations of latent clusters of the input graph.
\end{itemize}    

\section{Experimental Setups} \label{exp}
This section briefly presents the detailed training setups in the experiments.

\textbf{Implementation Details.} 
We implement our proposed \name based on the code of SAGPool~\cite{sagpool} and GMT~\cite{gmt}. Thus we use the same experimental setting. 
Specifically, for the model configuration of \name, we try the pooling ratio from $\{0.5, 0.25\}$, the hidden dimension from $\{128, 96, 64\}$ and the batch size from $\{64, 128\}$. Parameters are optimized by Adam optimizer with learning rate from $\{1 \times 10^{-3}, 5 \times 10^{-4}, 3 \times 10^{-4}\}$, weight decay from $\{2 \times 10^{-4}, 1 \times 10^{-4}\}$ and dropout rate from $\{0.5, 0.3\}$.
Furthermore, most reported results of baselines are derived from GMT, since GMT is the most recent work that replicates these popular pooling methods in previous studies. 
For ENZYMES, PTC-MR and Synthie, we adopt the recommended setting to perform parameter tuning based on the provided official code, including the source code of HaarPool~\cite{haarpool} and GMT~\cite{gmt}, and the code from PyTorch Geometric Library~\cite{pyg} for the rest of models. All experiments are conducted on a Linux server with 1 I9-9900k CPU, 1 RTX 2080TI GPU and 64G RAM.

\begin{table}[t]
  \caption{Statistics of the datasets. $\mathrm{V}_{avg}$ ($\mathrm{E}_{avg}$) represents the average number of nodes (edges), respectively.}
  \label{tab:stat}
  \begin{tabular}{lrrrr}
    \toprule
    Dataset & \# Graph & $\mathrm{V}_{avg}$ & $\mathrm{E}_{avg}$ & \# Class\\
    \midrule
    MUTAG & 188 & 17.93 & 19.79 & 2 \\
    ENZYMES & 600 & 32.63 & 124.20 & 6 \\
    PROTEINS & 1173 & 39.06 & 72.82 & 2 \\
    PTC-MR & 344 & 14.30 & 14.69 & 2 \\
    Synthie & 400 & 20.85 & 32.74 & 4 \\
    IMDB-BINARY & 1000 & 19.77 & 96.53 & 2 \\
    IMDB-MULTI & 1500 & 13.00 & 65.94 & 3 \\
    HIV & 41127 & 25.51 & 27.52 & 2 \\
    TOX21 & 7831 & 18.57 & 19.30 & 12 \\
    TOXCAST & 8576 & 18.78 & 19.30 & 617 \\
    BBBP & 2039 & 24.06 & 26.00 & 2 \\
  \bottomrule
\end{tabular}
\end{table}

\end{document}